\documentstyle[12pt,twoside,mkstyle]{article}
\newcommand{\Zitat}[1]{$^{\tiny\mbox{\cite{#1}}}$}

\date{}
\title{A Comment on "Analysis of Video Image Sequences Using Point and 
Line Correspondences"} 
\shorttitle{On "... Point and Line Correspondences"} 
\version{02}

\begin{document}
\setlength{\unitlength}{1cm}
\vfill
\machetitel
\thispagestyle{empty}
\vfill
\newpage
\begin{abstract}
In this paper we would like to deny the results of Wang et 
al.\Zitat{linehelps} raising two fundamental claims: 
\begin{itemize}
\item A line does not contribute anything to recognition of motion parameters 
from two images
\item Four traceable points are not sufficient to recover motion parameters 
from two perspective\footnote{Everywhere I replaced "prospective"
with "perspective" without further notifying this} 
 projections. 
\end{itemize} 
To be constructive, however, we show that four traceable points are 
sufficient 
 to recover motion parameters from two frames under orthogonal projection and 
that five points are sufficient to simplify the solution of the two-frame 
problem under orthogonal projection to solving a linear equation system.\\
\end{abstract}

\section{Introduction}
    Wang\Zitat{linehelps} claimed a new method 
of recovering structure and motion parameters from a sequence of two frames 
(under perspective projection) applicable whenever 
four points and a straight line form a rigid body and can be traced (or, in 
the languages of the 
authors: line and points correspondence is known) from frame to frame.  
We show here that the claim 
 of the Authors is wrong at least for two basic reasons:
\begin{itemize}
\item A line does not contribute anything to recognition of motion parameters 
from two images
\item Four traceable points are not sufficient to recover motion parameters 
from two perspective projections. 
\end{itemize} 
    To show that a line does not  
contribute anything to recognition of motion parameters from two images 
(Section~\ref{theline})
\begin{itemize}
\item we recall the results of Weng\Zitat{onlylines}
\item we consider degrees of freedom and amount of information connected with 
a line in each image
\item we point at errors in the proof of the results in~\Zitat{linehelps}
\end{itemize}
     To show, that four traceable points are not sufficient to recover motion 
parameters from two perspective projections (Section~\ref{fourprosp})
\begin{itemize}
\item we make a degrees of freedom argument
\item we demonstrate that infinitely many four point objects may be 
constructed from two images 
\item we explain errors in the proof given in ~\Zitat{linehelps}
\end{itemize}
    To complete the argument, we show that five points are sufficient for 
recovery of motion parameters from two perspective images 
(Section~\ref{fiveprosp}) \\
    To be constructive, however, we show that four traceable points are 
sufficient 
to recover motion parameters from two frames under orthogonal projection. 
We show also that one additional traceable point or one additional frame 
are sufficient to simplify the solution of the structure and motion recovery 
problem under orthogonal projection to solving a linear equation system 
(Section~\ref{ortho}).
\section{A Line Does Not Help The Points} \label{theline}
\subsection{The Argument After Weng} 
    In the paper of Weng\Zitat{onlylines} dealing with the recovery of motion 
and structure from line correspondences we find on page 319 the subsection 
{\it "A. Why Two Views Are Not Sufficient"}. The basic argument lies in the 
fact that the straight line we want to recover from perspective projections 
must lie in every plane containing the image of this line and the respective 
focal point. If we have two images, we have also two planes only to fit the 
recovered straight line which is almost always possible as two planes in 3-D 
in general intersect along a line. Though the argument of 
Weng\Zitat{onlylines} 
deals with any number of lines only, it is easily extended to the situation 
of Wang\Zitat{linehelps}: namely whenever points are met for two 
perspective images, any number of straight lines may be in general met. \\
\subsection{The Degrees of Freedom Argument}  \label{dofarg}
We can show the uselessness of a straight line for structure recovery from 
two images in a still another way. If we trace a rigid body through several 
(e.g. two) images we have the following degrees of freedom: 
the sum of degrees 
 of freedom for each element of the rigid body in the first frame and the 
degrees of freedom for motion for each subsequent frame (5 for orthogonal and 
6 for perspective projections if no information on constraints for motion 
are available). On the other hand each frame contributes some information on 
the positions of the traced elements of the rigid body. Now we can recover 
the structure of the rigid body if within each but the first frame the 
information gained outweighs the degrees of freedom introduced by the motion 
so that eventually it is possible to bind all the degrees of freedom of the 
first frame. We can reverse this formulation: the combined information gained 
on each traceable element over all frames should outweigh the degrees 
of freedom of this element emerging in the first frame so that superfluous 
information summed over all traced elements binds the degrees of freedom of 
the motion introduced by all but first frame. \\

    If we consider e.g. a point then we see that in the first frame it has 3 
degrees of freedom (df.) (e.g. x,y,z-co-ordinates). Each frame contributes 2 
independent pieces of information on a point (e.g. x,y-co-ordinates within 
the projection plane). Hence two projections bind 4 df. which leaves 1 
superfluous df. for recovery of motion parameters.\\

    If we consider a straight line, however, then in the first frame it has 4 
degrees of freedom (df.) (e.g. x,y-co-ordinates of the crossing point of the 
XY-plane, the angle between the XY-plane and the straight line, and the angle 
of rotation around the Z axis). Each frame contributes 
2 independent pieces of information on a straight line (e.g. 
a,b-parameters of the y=ax+b equation of the projection within 
the x,y-co-ordinate system of  
the projection plane). {\bf Hence two projections bind 4 df. which leaves NO 
superfluous df. for recovery of motion parameters, hence a traced straight 
line does not contribute anything to reconstruction of the rigid body from 
two frames. Q.e.D.}.
%
%
%
\subsection{Errors in The Proof of Wang} 
     Let us discuss now the contribution of Wang\Zitat{linehelps}, 
specifically with the proof in Section "2.3.Specifying constraints for 
mutually independent equations". On page 
1070 it is claimed that "Once the distance and angular constraints ... are 
imposed, the shape of points/line configuration is uniquely determined". It 
is not. Any mirror-like reflection of the configuration retains the angular 
and distance constraints but the shape is obviously different (mirror 
image).\\

      The most tricky step of the proof is step (5) (page 1069).
 While all the 
previous steps impose constraints selecting a point position out of a 
continuum, the step (5) simply means selection of a value out of a discrete 
set of (two) points. This fact should be paid a special attention when 
considering methods of solving the derived non-linear equation system
 (Section 3, page 1071). It is obvious that if the initial guess is close 
enough to the wrong of the two discrete values then the methods solving 
non-linear systems may make hardly any use of the criterion from step (5).  
%
%
\section{The Set Of Four Point Bodies Recovered From Two Prospective 
Projections} \label{fourprosp}
    As we have shown in the previous section, a line is not helpful for 
recovering the shape and motion parameters of a rigid body from two frames 
under perspective projection. So either four points are sufficient for 
recovery or the 
results of Wang\Zitat{linehelps} are geometrically wrong. We claim that the 
latter is the case.
\subsection{Degrees of Freedom Argument}
    Let us have a closer look at degrees of freedom than in 
Section~\ref{dofarg} of this paper. In general, with p points and s straight 
lines forming the rigid body traced over k frames we have the 
following number 
of degrees of freedom:
$$-1+3*p+4*s+6*(k-1)$$
The constituent -1 is due to the fact that the scaling of the object 
cannot 
be recovered under perspective projection. The factor 3 means the number of 
degrees of freedom for a point, 4 - for a straight line and 
6 - for the motion between frames.\\

    Now the amount of information gained within those k frames amounts to: 
$$k*(2*p+2*s)$$
    In order to recover the structural and motion data we request that:
\begin{equation}
                -1+3*p+4*s+6*(k-1) \le k*(2*p+2*s)
\end{equation}  
    When we have to do with 2 frames (k=2) and 4 points (p=4, s=0) only, 
we obtain:.
$$-1+3*p+4*s+6*(k-1)=-1+12+6=17  > k*(2*p+2*s)=2*2*4=16$$  
\noindent which means that Wang\Zitat{linehelps}  has missed the solution by 
exactly one degree of freedom. \\

    Let us notice, however, that with 3 frames 
(k=3) and 4 points (p=4, s=0) we obtain
$$-1+3*p+4*s+6*(k-1)=-1+12+12=23 < k*(2*p+2*s)=3*2*4=24$$  
\noindent ensuring the existence of a solution (see K{\l}opotek\Zitat{K:3}).\\
    Also with 2 frames 
(k=2) and 5 points (p=5, s=0) we obtain
$$-1+3*p+4*s+6*(k-1)=-1+15+6=20 = k*(2*p+2*s)=2*2*5=20$$  
\noindent ensuring the existence of a solution 
(see Nagel\Zitat{fivepoints}).\\
    Also with 3 frames 
(k=3) and 3 points and a single line (p=3, s=1) we obtain
$$-1+3*p+4*s+6*(k-1)=-1+9+4+12=24 = k*(2*p+2*s)=3*(6+2)=24$$  
\noindent ensuring the existence of a solution.\\
    With 3 frames 
(k=3) and six lines (p=0, s=6) we obtain
$$-1+3*p+4*s+6*(k-1)=-1+0+24+12=35 < k*(2*p+2*s)=3*(0+12)=36$$  
\noindent ensuring the existence of a solution (compare 
Weng\Zitat{onlylines}).\\

    The subsequent section points at the failure of Wang\Zitat{linehelps} 
more 
explicitly by demonstrating that it is possible to construct (uncountably 
infinitely) many objects fitting given two projection frames.\\

\input 3-5pic.tex
 \subsection{Objects Recoverable from Two Frames} \label{construction}
    Let us imagine now we have two perspective projections of a four point 
body. Instead of considering the points A',B',C',D' and A",B",C",D" being 
projections of the four points A,B,C,D in the first and the second 
projection, 
respectively and the focal points $f_1$ and $f_2$ let us consider the straight
lines a'=$f_1$A' (=$f_1$A), b'=$f_1$B' (=$f_1$B), c'=$f_1$C' (=$f_1$C),
d'=$f_1$D' (=$f_1$D),
a"=$f_2$A" (=$f_2$A), b"=$f_2$B" (=$f_2$B), c"=$f_2$C" (=$f_2$C), d"=$f_2$D"
(=$f_2$D), assuming at the 
same time that the camera was moving rather than the four point body itself.
(See scene in Fig.1
\footnote{Figures have been added, and the content of this subsection has
been changed to utilize images}
 - REMARK:
All the Figures 1-8 are orthogobnal projections of an imaginary scene, where 
the coordinate system origin is right in the center of the cube drawn in each
image.  Each of figures 1-8 consists of two images: the
first one was taken uniformly from the same projection angle to achieve cross
reference among figures. The second image represents the same scene but from a
different side, that side which seemed most reasonable for the author of this
paper). Obviously, our (fixed) observables are the angles A$f_1$B,
 A$f_1$C, A$f_1$D,
 B$f_1$C, B$f_1$D, C$f_1$D,  A$f_2$B, A$f_2$C, A$f_2$D, B$f_2$C, B$f_2$D, 
C$f_2$D, \\

    We can assume (due to the unknown scaling) that we know the length of 
the 
edge AB. We fix further the plane p' containing A,B and $f_1$ (as the
z=0 plane in the figures). Within this plane 
the point $f_1$ can be any point  on a circle, determined by the size of
the (fixed) angle
A$f_1$B and location of two points A,B belonging to it - see Fig.2. 
(Strictly speaking it lies on one of the two circle fragments cut out by
points A and B - this is a well knowwn fact from elementary geometry).
 Let us select and fix one of those candidate points for $f_1$. \\

    Under these circumstances the point $f_2$ belongs to a rotational surface
- with line AB being the rotational axis - 
containing all the circles determined by the size of 
the (fixed) angle A$f_2$B and location of points A,B belonging to all of them 
(see Fig.2, for imagination of geometric place of $f_2$). The degrees of
freedom for positioning $f_1$ on this rotational body are indicated in Figs
3 and 4. Fig.3 shows the surface drawn in space by the line c" if we fix
the angle  $f_2$AB and let rotate $f_2$ around the
AB axis. Fig.4 shows aa fragment of surface drawn in space by the line c"
if we fix an $f_2$AB   circle and  let  $f_2$  move around this circle.\\

 Let us concentrate now on identification of points A,B,C in space
that is after fixing position of A and B in space we try to
 find  the geometrical place of crossing of straight lines
c' 
and c" (Fig.5) when $f_1$ is fixed and $f_2$ can move freely (up to
geometrical constraints indicated above). 
If we select an angle  $f_2$AB   and let $f_2$ rotate
around the AB axis then
a paraboloidal surface is drawn by straight line c" as indicated in Fig.3, 
so that the straight line c' in general has a chance crossing this surface,
just indicating the position of point C in space. 
Then in general, due to continuity, 
 hen c'  crosses also in the neighbourhood the 
surface drawn by 
c" for a neighbouring angle $f_2$AB. 
 So for a continuous set of angles $f_2$AB there exists a (continuous)
curve r such that for any point $f_2$ lying on it the lines c' and c" cross
one another. If we let $f_2$ move along this curve r then c" will draw in
space a surface indicated in Fig.5. \\

 Now let us consider the possibility of lines d' and d" crossing each 
other when we have fixed the point $f_1$. When the point $f_2$ runs along the
curve r, the straight line d" draws
a surface in space (Fig.6). Let us assume that for the current position of
point
$f_1$ the line d' crosses in fact this surface  (Fig.6). Then we have matched
the two images into 
a four point 3-D body.\\

    However, let us investigate neighbouring positions of the point $f_1$ on
its circle in the plane p'  (see Fig.7 and 8) - that is we do not assume
fixed position of $f_1$ anymore.  We see with ease that in general case
we can meet
 all the four pairs of lines a' and a", b' and b", c' and c", d' and d"
also in the vicinity of the
original matching - compare Fig.5 with 7, and 6 with 8 - because
for a slightly shifted point $F_1$ the curve r for geometrical place of $f_2$
for meeting three line pairs (a'-a",b'-b",c'-c") will move slightly in space,
so will the surface drawn by c" on curve r (see Fig.7), and so will the
surface drawn by d" (see Fig.8). Therefore, the slightly moved d' will in
general have the chance to hit also the slightly moved surface of d". \\

Hence in fact {\bf there exists an unlimited set
of possible
"recovered" four point bodies corresponding to given two perspective images.}
\\

    If we combine these results with the results of section~\ref{theline}, 
we see that structure and motion cannot be uniquely determined from the data
 exploited in~\Zitat{linehelps}. The fact of obtaining some experimental 
results 
in~\Zitat{linehelps} can be explained simply by the nature of approximate 
algorithms used there in combination with a bit complicated spatial structure 
of the set of four point bodies corresponding to the given two 
projection frames. 
\subsection{Fundamental Errors in the Proof Given by Wang} 
    Let us confront the above results with the (length and angle) invariant 
method applied in the paper~\Zitat{linehelps}. Clearly the estimates of the 
number of degrees of freedom and of information content will be different
 from 
those exploited in the previous subsections. However, the balance between 
them will remain the same. As stated in~\Zitat{linehelps}, the four points 
deliver 6 independent invariants (lengths of line segments joining each pair 
of points). The number of variables introduced by the invariant method is 
equal 8 (two for each point, as there are two frames). One unknown may be 
bound by the non-determinability of the scale of the traced object. So 
we have 
6 equations in 7 independent variables which is obviously unsolvable. \\

    Now let us look at the introduction of the line (whether we consider 
first the line, then the points or vice versa should not change the estimates 
of the number of unknowns and the number of equations). Obviously, as stated 
in~\Zitat{linehelps}, 4 new variables are introduced this way. So we would
 need 
 5 more independent 
equations (5 more invariants) to reach a solvable equation 
 system. Given 4 non-coplanar points in 3-D space, we can span a coordinate 
system within this space. It is easily shown, that at least two triples of 
points out of those four span planes cut by the straight line. To determine 
the cutting point within each plane we need two distances - from two of the 
plane spanning points. Hence, four independent invariants are introduced 
instead of five which are needed. So the reconstruction process has to fail. 
\\

    At this point we shall make the remark that the above-mentioned four 
distances do not determine the position of the line uniquely, as distances 
from two points in a plane determine not a single, but rather 
two points lying 
symmetrically, so we get in fact four straight lines out of the 
distance invariants mentioned above. To get a unique line, a fifth distance 
may be required (e.g. of the added line to one of the four points). However, 
this distance would not be independent of the other ones. The firs four 
distances may be selected out of a continuum (in such a way as to let 
respective circles cross each other), but the fifth distance may obtain only 
values from a discrete set of four values. \\
However, we do not need to restrict ourselves to distance-like invariants. In 
fact, if we bind to the four points a coordinate system spanned by them then 
the four real-valued coordinates of the straight line within 
this coordinate system are the invariants uniquely determining the position 
of the line. Hence only four truly independent equations may be established 
which contradicts the argument and the formula of Wang given in 
Section 2.4. on 

page 1070\Zitat{linehelps}. (Wang 
claims there that a line introduces six independent equations as it were  
uniquely defined by 6 constraints). 
\section{Recovery With Five Traceable Points And Two Prospective Images} 
\label{fiveprosp}
    It has already been proposed to use 5 traceable points to recover 
structure and motion parameters from two perspective 
images\Zitat{fivepoints} (compare also Roach and \linebreak
Aggarwal\Zitat{R:1}~\Zitat{R:2}). 
Let us propose here another construction of an equation system doing the 
job, based on the construction described in section~\ref{construction} of this

paper.\\
 
   Let us 
imagine now we have two perspective projections of a five point 
body. Instead of considering the points A',B',C',D',E' and A",B",C",D",E" 
being 
projections of the five points A,B,C,D,E in the first and the second 
projection, 
respectively and the focal points $f_1$ and $f_2$ let us consider the straight
lines a'=$f_1$A' (=$f_1$A), b'=$f_1$B' (=$f_1$B), c'=$f_1$C' (=$f_1$C),
d'=$f_1$D' (=$f_1$D), e'=$f_1$E' (=$f_1$E), 
a"=$f_2$A" (=$f_2$A), b"=$f_2$B" (=$f_2$B), c"=$f_2$C" (=$f_2$C), d"=$f_2$D"
(=$f_2$D), e"=$f_2$E" (=$f_2$E), assuming at the 
same time that the camera was moving rather than the four point body itself.\\

    We can assume (due to the unknown scaling) that we know the length of 
the 
edge AB. We fix further the plane p' containing A,B and $f_1$. Within this
plane the point $f_1$ lies on a circle (determined by the size of the angle 
A$f_1$B) containing the points A,B. 
    Under these circumstances the point $f_2$ belongs to a rotational surface 
containing all the circles containing A and B and determined by the size of 
the angle A$f_2$B. \\

    The position of 
the focal point $f_1$ may be characterized by the angle AB$f_1$, the position
of the focal point $f_2$ may be characterized by the angle AB$f_2$ and by the 
rotational angle of the AB-axis (3 variables). We shall require that the 
straight lines c',d',e' meet with c",d",e" respectively. Analytically, each 
of the lines is characterized by two vectors (determined from the variables 
previously described) and a freely ranging parameter ($p+t*v$), so the six 
lines introduce 6 variables. They provide us with 9 equations (3 for each 
dimension of any meeting line pair). So we obtain an equation system with 9 
equations in 9 variables.
%

\section{On The Recovery Under Orthogonal Projection} \label{ortho}
 
    It is an interesting question to investigate the possibility of 
reconstruction of structure and motion from multiframes under orthogonal 
projection. As mentioned in~\Zitat{K:3}, it is possible to recover them from 
three traceable points and three images having a quadratic equation system, 
which may be simplified to a linear one if four images are available. This 
section will exploit those results to show the possibility of reconstruction 
from two images when four points are traceable and the possibility of 
simplification to linear equation systems whenever either five points are 
traceable (instead of four) or three images are available (instead of two). 
The respective subsections will be preceded by a degree of freedom 
consideration. 
\subsection{Degrees of Freedom for Orthogonal Projection}
    As in case of perspective projections, each point introduces 3 df in the 
first frame, each line - 4 df minus one df for the whole body as there exists 
no possibility of determining the initial depth of the
 body in the space. The 
motion introduces for each subsequent 
frame 5 df only, because the motion in the direction orthogonal to the 
projection plane has no impact on the image. In general, with p points and s 
straight 
lines forming the rigid body traced over k frames we have 
$$-1+3*p+4*s+5*(k-1)$$ degrees of freedom against
$$ k*(2*p+2*s)$$ pieces of information available from k images.\\
    Thus we shall have the balance 
\begin{equation}
                -1+3*p+4*s+5*(k-1) \le  k*(2*p+2*s)
\end{equation}
to achieve recoverability. \\

    Let us consider some combinations of parameters:\\
\begin{itemize}
\item for k=3 frames, p=3 points we get
$$-1+3*p+4*s+5*(k-1)=18 = k*(2*p+2*s)=18$$ 
\item for k=2 frames, p=4 points we get
$$-1+3*p+4*s+5*(k-1)=-1+12+5=16 =  k*(2*p+2*s)=2*2*4=16$$ 
\end{itemize}
On exploiting\footnote{Additional information} straight line component of
the above equation see\Zitat{K:4}, and on non-geometrical balancing degrees
of freedom see\Zitat{K:2}. 
\subsection{Structure and Motion for 4 Point Correspondences}
    K{\l}opotek\Zitat{K:2} presented a method for recovery of
structure and motions 
parameters for 3 traceable points and 3 frames under orthogonal projection. 
Let us briefly review the results as they form a basis for 
considerations of this section. \\

    Let A,B,C be the traced points of a rigid body, and $A_i,B_I,C_i$ their 
respective projections within the ${i}^{th}$ frame. Let $a,b,c,a_i,b_i,c_i$ 
denote the lengths of straight line segments 
$BC,AC,AB,B_iC_i,A_iC_i,A_iB_i$, respectively. Then for each frame one of 
the following relationships holds:\\
Either
$$ \sqrt{{a}^2-{a_i}^2}+\sqrt{{b}^2-{b_i}^2}-\sqrt{{c}^2-{c_i}^2}=0$$
or
\begin{equation} \label{p3f3}
 \sqrt{{a}^2-{a_i}^2}-\sqrt{{b}^2-{b_i}^2}+\sqrt{{c}^2-{c_i}^2}=0
\end{equation}
or
$$-\sqrt{{a}^2-{a_i}^2}+\sqrt{{b}^2-{b_i}^2}+\sqrt{{c}^2-{c_i}^2}=0$$
(which is easily seen from geometrical relationships, 
presented analytically and graphically\footnote{Additional
information} by K{\l}opotek\Zitat{Klopotek:92g}).
 So
we have three
equations, for i=1,2, and 3, in three unknowns, a,b,c. As any of the above 
relationships gives after a twofold squaring:
\begin{eqnarray}
{a}^4+{b}^4+{c}^4-2{a}^2{b}^2-2{a}^2{c}^2-2{b}^2{c}^2+\nonumber\\
{a_i}^4+{b_i}^4+{c_i}^4-2{a_i}^2{b_i}^2-2{a_i}^2{c_i}^2-2{b_i}^2{c_i}^2=
\nonumber\\
=2{a}^2( {a_i}^2-{b_i}^2-{c_i}^2)+
   2{b}^2(-{a_i}^2+{b_i}^2-{c_i}^2)+ \label{p3f3sqr}\\
   2{c}^2(-{a_i}^2-{b_i}^2+{c_i}^2)\nonumber
\end{eqnarray}
which is quadratic in ${a}^2,{b}^2,{c}^2$, hence solvable by exploitation of 
respective methods. \\

    So let us consider a rigid body with four points over two frames (i=1,2). 
\\
    Let $A$, $B$, $C$, $D$ be the traced points of a rigid body, and 
$A_i$, $B_I$, $C_i$, $D_i$ their 
respective projections within the ${i}^{th}$ frame. Let $a$, $b$, $c$, $d$, 
$e$, $f$, $a_i$, $b_i$, $c_i$, $d_i$, $e_i$, $f_i$ 
denote the lengths of straight line segments 
$BC$, $AC$, $AB$, $AD$, $BD$, $CD$, $B_iC_i$,  
$A_iC_i$, $A_iB_i$, $A_iD_i$, $B_iD_i$, $C_iD_i$, 
respectively. Then for each frame three relationships hold:\\
$$ \sqrt{{e}^2-{e_i}^2}=\sqrt{{a}^2-{a_i}^2}+\sqrt{{f}^2-{f_i}^2}$$
and
\begin{equation} \label{p4f2}
 \sqrt{{d}^2-{d_i}^2}=\sqrt{{b}^2-{b_i}^2}+\sqrt{{f}^2-{f_i}^2}
\end{equation}
and
$$ \sqrt{{e}^2-{e_i}^2}=\sqrt{{c}^2-{c_i}^2}+\sqrt{{d}^2-{d_i}^2}$$
or another proper variation according 
to the possibilities mentioned for three 
 points.\\
   Please notice that we have also a fourth relationship related to the 
triangle ABC:\\
$$\sqrt{{a}^2-{a_i}^2}=\sqrt{{b}^2-{b_i}^2}+\sqrt{{c}^2-{c_i}^2}$$\\
but we make no use of it as it is linearly dependent on the three previous 
ones.\\

    In this way we got 6 equations (3 for each of the two frames) in six 
variables a,b,c,d,e,f. The respective twofold squaring leads to quadratic 
equations.
\subsection{Linearization of the Equation System} \label{lineariz}
    In the opinion of the author of this paper it is easier to solve a linear 
equation system than a quadratic one, especially if no good guess values are 
available to start an iteration with. For this reason he considers it to be a 
good practice to exploit any available redundancy to make the 
problem a linear one and 
only to exploit the redundancy for noise reduction after finding a first 
satisfying approximation. He is not isolated in this thinking as may be seen 
from Weng\Zitat{onlylines} where 13 lines were traced (instead of sufficing 
6).\\

    K{\l}opotek\Zitat{Klopotek:92g} simplified the equation system
(\ref{p3f3sqr}) for 3 
traceable points by 
using four instead of three frames and subtracting the twofold squared 
equation for the first 
frame from those of the other ones. So one obtains three equations of the 
form for i=2,3,4: 
\begin{eqnarray}
 {a_i}^4+{b_i}^4+{c_i}^4-2{a_i}^2{b_i}^2-2{a_i}^2{c_i}^2-2{b_i}^2{c_i}^2-
\nonumber\\
{a_1}^4-{b_1}^4-{c_1}^4+2{a_1}^2{b_1}^2+2{a_1}^2{c_1}^2+2{b_1}^2{c_1}^2=
\nonumber\\
=2{a}^2( {a_i}^2-{b_i}^2-{c_i}^2-{a_1}^2+{b_1}^2+{c_1}^2)+\\
 2{b}^2(-{a_i}^2+{b_i}^2-{c_i}^2+{a_1}^2-{b_1}^2+{c_1}^2)+ \nonumber\\
 2{c}^2(-{a_i}^2-{b_i}^2+{c_i}^2+{a_1}^2+{b_1}^2-{c_1}^2) \nonumber
\end{eqnarray}
which are linear in  ${a}^2,{b}^2,{c}^2$, hence solvable by exploitation of 
respective methods. (No linear dependence is introduced as a new frame is 
exploited unless the motion has a very special form).\\

    The very same approach may be used with the four points: instead of 2 
frames we take one more. Then from (\ref{p4f2})we obtain the linear equation 
system for i=2,3:\\
\begin{eqnarray}
{a_i}^4+{e_i}^4+{f_i}^4-2{a_i}^2{e_i}^2-2{a_i}^2{f_i}^2-2{e_i}^2{f_i}^2-
\nonumber\\
{a_1}^4-{e_1}^4-{f_1}^4+2{a_1}^2{e_1}^2+2{a_1}^2{f_1}^2+2{e_1}^2{f_1}^2=
\nonumber\\
 =2{a}^2( {a_i}^2-{e_i}^2-{f_i}^2-{a_1}^2+{e_1}^2+{f_1}^2)+
\nonumber\\
   2{e}^2(-{a_i}^2+{e_i}^2-{f_i}^2+{a_1}^2-{e_1}^2+{f_1}^2)+ 
\nonumber\\
 2{f}^2(-{a_i}^2-{e_i}^2+{f_i}^2+{a_1}^2+{e_1}^2-{f_1}^2) 
\nonumber
\end{eqnarray}
and
\begin{eqnarray}
 {d_i}^4+{b_i}^4+{f_i}^4-2{d_i}^2{b_i}^2-2{d_i}^2{f_i}^2-2{b_i}^2{f_i}^2-
\nonumber\\
{d_1}^4-{b_1}^4-{f_1}^4+2{d_1}^2{b_1}^2+2{d_1}^2{f_1}^2+2{b_1}^2{f_1}^2=
\nonumber \\
  =2{d}^2( {d_i}^2-{b_i}^2-{f_i}^2-{d_1}^2+{b_1}^2+{f_1}^2)+\\
   2{b}^2(-{d_i}^2+{b_i}^2-{f_i}^2+{d_1}^2-{b_1}^2+{f_1}^2)+ \nonumber\\
   2{f}^2(-{d_i}^2-{b_i}^2+{f_i}^2+{d_1}^2+{b_1}^2-{f_1}^2) \nonumber
\end{eqnarray}
and
\begin{eqnarray}
{d_i}^4+{e_i}^4+{c_i}^4-2{d_i}^2{e_i}^2-2{d_i}^2{c_i}^2-2{e_i}^2{c_i}^2-
\nonumber\\
{d_1}^4-{e_1}^4-{c_1}^4+2{d_1}^2{e_1}^2+2{d_1}^2{c_1}^2+2{e_1}^2{c_1}^2=
\nonumber\\
=2{d}^2( {d_i}^2-{e_i}^2-{c_i}^2-{d_1}^2+{e_1}^2+{c_1}^2)+
\nonumber\\
   2{e}^2(-{d_i}^2+{e_i}^2-{c_i}^2+{d_1}^2-{e_1}^2+{c_1}^2)+ 
\nonumber\\
 2{c}^2(-{d_i}^2-{e_i}^2+{c_i}^2+{d_1}^2+{e_1}^2-{c_1}^2) 
\nonumber
\end{eqnarray}
\\
This linear equation system is easily solved.\\

    Another possibility of linearization is to stay with 2 frames but
 to use 5 
traceable points instead of four. With 5 traceable points A,B,C,D,E we are 
interested in recovering 9 line segments: $AB$, $AC$, $BC$, 
$AD$, $BD$, $CD$, $AE$, $BE$, $CE$. The 
tenth line segment DE is also of interest, though available from the other 
ones. Let us turn to twofold squared equations for each of the 10 triangles 
$ABC$, $ABD$, $ACD$, $BCD$, $ABE$, $ACE$, $BCE$, $ADE$, $BDE$, $CDE$. For the 
triangle ABC we have:
\begin{eqnarray}
 {BC_2}^4+{AC_2}^4+{AB_2}^4-2{BC_2}^2{AC_2}^2
-2{BC_2}^2{AB_2}^2-2{AC_2}^2{AB_2}^2- \nonumber\\
 {BC_1}^4-{AC_1}^4-{AB_1}^4+2{BC_1}^2{AC_1}^2
+2{BC_1}^2{AB_1}^2+2{AC_1}^2{AB_1}^2= \nonumber\\
 =2{BC}^2( {BC_2}^2-{AC_2}^2-{AB_2}^2
-{BC_1}^2+{AC_1}^2+{AB_1}^2)+         \\
 2{AC}^2(-{BC_2}^2+{AC_2}^2-{AB_2}^2
+{BC_1}^2-{AC_1}^2+{AB_1}^2)+         \nonumber\\
 2{AB}^2(-{BC_2}^2-{AC_2}^2+{AB_2}^2
+{BC_1}^2+{AC_1}^2-{AB_1}^2)\nonumber
\end{eqnarray}
\noindent and so on for each of the ten triangles. Notice that though the 
not squared equations for each frame are linearly dependent, the two-fold 
squared subtracted equations are not linearly dependent. Hence the equation 
system with ten linear equations in 10 unknown squared line segment lengths  
${AB}^2$, ${AC}^2$, ${BC}^2$, 
${AD}^2$, ${BD}^2$, ${CD}^2$, ${AE}^2$, ${BE}^2$, ${CE}^2$, ${DE}^2$ is 
solvable.
%

\section{Discussion and Conclusions}
    This paper was dealing with the problem of recovery of 3-D structure and 
motion from multiframes. In the past, this problem has been discussed under 
various 
geometrical and physical settings. Roach and Aggarwal~\Zitat{R:1}~\Zitat{R:2} 
researched on motion and structure recovery tracing points under perspective 
projection assuming static 
scene and moving camera. They showed that five points in two views are needed 
to recover the structure and motion parameters. Their solution involved a 
system of 18 highly nonlinear equations. Nagel~\Zitat{fivepoints} proposed a 
simplified equation system by separating solution for the translation vector 
and the rotation matrix, with rotation matrix being determined by a system of 
3 equations in three motion parameters. Weng et al~\Zitat{onlylines}
 recovered 
 structure and motion from line correspondences only using three frames and 13
lines, obtaining a linear equation system solving the problem. \\

   Simpler solutions of the problem of 
recovery of structure and motion from multiframes are achievable under 
orthogonal projections and under special assumptions concerning the physical 
motion, which allow for reduction of the quantity and quality  of required 
traceable features 
 and/or the complexity of the solution.
 Lee~\Zitat{Lee:1} recovered structure 
 and motion tracing two 
points only under orthogonal projection from four frames assuming constant 
rotational speed of the traced body. K{\l}opotek~\Zitat{K:2} also dealt with 
the 
case of two traceable points, assuming free fall of  the  body  in 
homogeneous 
force field. \\

    For other 
related papers see \cite{K:4}-\cite{Liu}.
      \\

    The paper of Wang~\Zitat{linehelps} represents an attempt to press down 
further the  quantity and quality  of required  traceable features under 
perspective 
projection. Wang attempts to use only four points and a line (instead of 
the earlier achievement of five points) over two frames. A line introduces 
generally more degrees of freedom (4 against 3 introduced by a point) and its 
position may be traced more reliably than that of a 
point~\Zitat{onlylines}. The gain would be obvious: the less features are 
required as a minimum the more stability may be pressed out of the actually 
available redundancy.\\

    However, the current paper points at a basic weakness of the Wang's 
experimentally widely elaborated paper: the claim of the possibility of the 
recovery of structure and motion from the claimed minimal set of features 
does not hold. The following arguments were presented: a) a line does not 
contribute anything to recovery from two frames only, b) four points alone 
do not suffice to recover structure and motion. This has been shown by a) 
evaluating the number of degrees of freedom against the amount of information 
obtainable from the frames, b) by constructing an infinite set of candidate 
rigid bodies giving two predefined projections. \\

    We have also pointed at the 
errors in the paper of Wang which led to those incorrect results. Three 
claims of Wang are wrong. The first error is a cosmetic one: Wang claims that 
a (sufficiently big) set of distance and/or angular constraints determines 
uniquely the shape of a rigid body consisting of points and lines. In reality 
it is not a full uniqueness  but  a  uniqueness  up  to  a  mirror 
symmetry. The 
 basic error is a serious one: Wang claims that a constraint selecting 
a value out of a countable finite set of values is independent of those 
constraints which led to such a restriction of the set of values. The second 
wrong claim is in some sense a consequence of the first one: if symmetries 
could be fully eliminated by adding constraints then these constraints were 
independent. The third error just summarizes the previous ones: Wang missed 
the possibility of defining invariants in such a way as to eliminate 
ambiguities resulting from symmetries: he should have substituted distances
 by 
 coordinates in a coordinate system bound to the recovered rigid body.\\

    This paper shows also that under orthogonal projection the four points 
would suffice with two frames to recover structure and motion. Furthermore, 
if one more point were available or the body were traced over one more frame 
then a linear equation system may be constructed to find the structure and 
motion parameters.\\

    Last not least I want to apologize to the Authors of the 
paper~\Zitat{linehelps}. The critical remarks made here should not undermine 
the value of the labour behind the paper nor the value of large portions  
of the 
results published because much of the simulation and experimental work has 
been done with a number of traceable points sufficient for recovery purposes. 
Hence practical conclusions and hints from that research work remain valid to 
a large extent.
%
%
%
%
\newpage
%
{\bf Summary:}
{\small
    Wang et al.\Zitat{linehelps} claimed a new method 
of recovering structure and motion parameters from a sequence of two frames 
(under perspective projection) applicable whenever correspondences for 
four points and a straight line belonging to a rigid body over two frames 
may be established..In this paper we would like 
to deny the results of Wang\Zitat{linehelps} 
raising two fundamental claims:
\begin{itemize}
\item A line does not contribute anything to recognition of motion parameters 
from two images because:
\begin{itemize}
\item from two images we obtain only as much information as to position a 
line relatively to other, earlier recovered objects,
\item three projections of a line are necessary to extract from them some 
contribution to recover structure and motion parameters of a rigid body.
\end{itemize}
\item Four traceable points are not sufficient to recover motion parameters 
from two perspective projections, because: 
\begin{itemize}
\item the number of degrees of freedom connected with the four point rigid 
body and the motion between the frames exceeds the amount of information that 
can be extracted from a pair of images,
\item a pair of images may correspond to an infinite number of rigid four 
point bodies
\end{itemize} 
\end{itemize} 
    We point at basic errors in the theoretical part of Wang's paper and try 
to explain briefly the validity of experimental results.\\
We show also that four traceable points are 
sufficient 
 to recover motion parameters from two frames under orthogonal projection and 
that five traceable points in two frames or four traceable points in three 
frames are sufficient to simplify the solution of the reconstruction
problem under orthogonal projection to solving a linear equation system.
}\\
\newpage
 
{\normalsize {\bf About the Author--} M.A.K{\l}opotek was born in Brusy, 
Poland, 
in 1960. He received his Master's (in 1983) and Ph.D. (in 1984) degrees in 
Information 
Processing from the University of Technology in Dresden, Germany. Since 1985 
he works in the Institute of Computer Science of Polish Academy of Sciences in

Warsaw. His research interests are in computer vision, artificial 
intelligence and creative data analysis. 
}

\end{document}